\documentclass[10pt,twocolumn,letterpaper]{article}
\usepackage{graphicx}
\usepackage{subcaption}
\usepackage{float}
\usepackage[justification=raggedright]{caption}	
\usepackage{lscape}                                         

\usepackage[lined,ruled,linesnumbered]{algorithm2e}

\usepackage{booktabs}                   
\usepackage{multirow}

\usepackage{paralist}

\usepackage{bm}                          
\usepackage{epsfig}                      
\usepackage{graphicx}                  
\usepackage{times}
\usepackage{mathptmx}
\usepackage{mathtools}
\usepackage{amssymb,amsmath}   

\usepackage{units}
\usepackage{color}

\usepackage{comment}

\usepackage{url}  

\usepackage{xspace}
\usepackage[table]{xcolor}
\usepackage{setspace}

\def\eg{e.g.,~}               

\newlength\paramarginsize
\newlength\figmarginsize
\newlength\secmarginsize
\newlength\figcapmarginsize
\newlength\tabcapmarginsize

\newcommand{\secref}[1]{Section~\ref{sec:#1}}
\newcommand{\figref}[1]{Figure~\ref{fig:#1}} 
\newcommand{\tabref}[1]{Table~\ref{tab:#1}}
\newcommand{\eqnref}[1]{\eqref{eq:#1}}

\newcommand{\appenref}[1]{Appendix \S ~\ref{appen:#1}}
\long\def\ignorethis#1{}

\def\ours{\texttt{PCBEAR}}

\newcommand{\best}[1]{{\textbf{#1}}}

\def\xi{\mathbf{x}_i}

\graphicspath{{figure}, {images}, {example}}

\usepackage{cvpr}{}      

\definecolor{cvprblue}{rgb}{0.21,0.49,0.74}
\usepackage[pagebackref,breaklinks,colorlinks]{hyperref}
\usepackage[accsupp]{axessibility}
\def\paperID{5} 
\def\confName{CVPRW}
\def\confYear{2025}

\begin{document}

\title{PCBEAR: Pose Concept Bottleneck for Explainable Action Recognition}

\author{
Jongseo Lee$^1$ \quad
Wooil Lee$^1$ \quad
Gyeong-Moon Park$^2$\textsuperscript{†} \quad
Seong Tae Kim$^1$\textsuperscript{\dag} \quad
Jinwoo Choi$^1$\textsuperscript{\dag}\\
$^1$Kyung Hee University, Republic of Korea \\
$^2$Korea University, Republic of Korea \\
{\tt\small \{jong980812, lwi2765, st.kim, jinwoochoi\}@khu.ac.kr, gm-park@korea.ac.kr}
}
\twocolumn[{
\renewcommand\twocolumn[1][]{#1}
\maketitle
}]
\renewcommand{\thefootnote}{\fnsymbol{footnote}}
\footnotetext[2]{Corresponding authors.}
\begin{abstract}
Human action recognition (HAR) has achieved impressive results with deep learning models, but their decision-making process remains opaque due to their black-box nature. Ensuring interpretability is crucial, especially for real-world applications requiring transparency and accountability. Existing video XAI methods primarily rely on feature attribution or static textual concepts, both of which struggle to capture motion dynamics and temporal dependencies essential for action understanding. To address these challenges, we propose \textbf{P}ose \textbf{C}oncept \textbf{B}ottleneck for \textbf{E}xplainable \textbf{A}ction \textbf{R}ecognition (PCBEAR), a novel concept bottleneck framework that introduces human pose sequences as motion-aware, structured concepts for video action recognition. Unlike methods based on pixel-level features or static textual descriptions, PCBEAR leverages human skeleton poses, which focus solely on body movements, providing robust and interpretable explanations of motion dynamics. We define two types of pose-based concepts: static pose concepts for spatial configurations at individual frames, and dynamic pose concepts for motion patterns across multiple frames. To construct these concepts, PCBEAR applies clustering to video pose sequences, allowing for automatic discovery of meaningful concepts without manual annotation. We validate PCBEAR on KTH, Penn-Action, and HAA500, showing that it achieves high classification performance while offering interpretable, motion-driven explanations. Our method provides both strong predictive performance and human-understandable insights into the model’s reasoning process, enabling test-time interventions for debugging and improving model behavior.
\end{abstract}

\vspace{-1.em}
\section{Introduction}
\label{sec:intro}
\vspace{-0.5em}

\begin{figure}
    \centering
    \includegraphics[width=\linewidth]{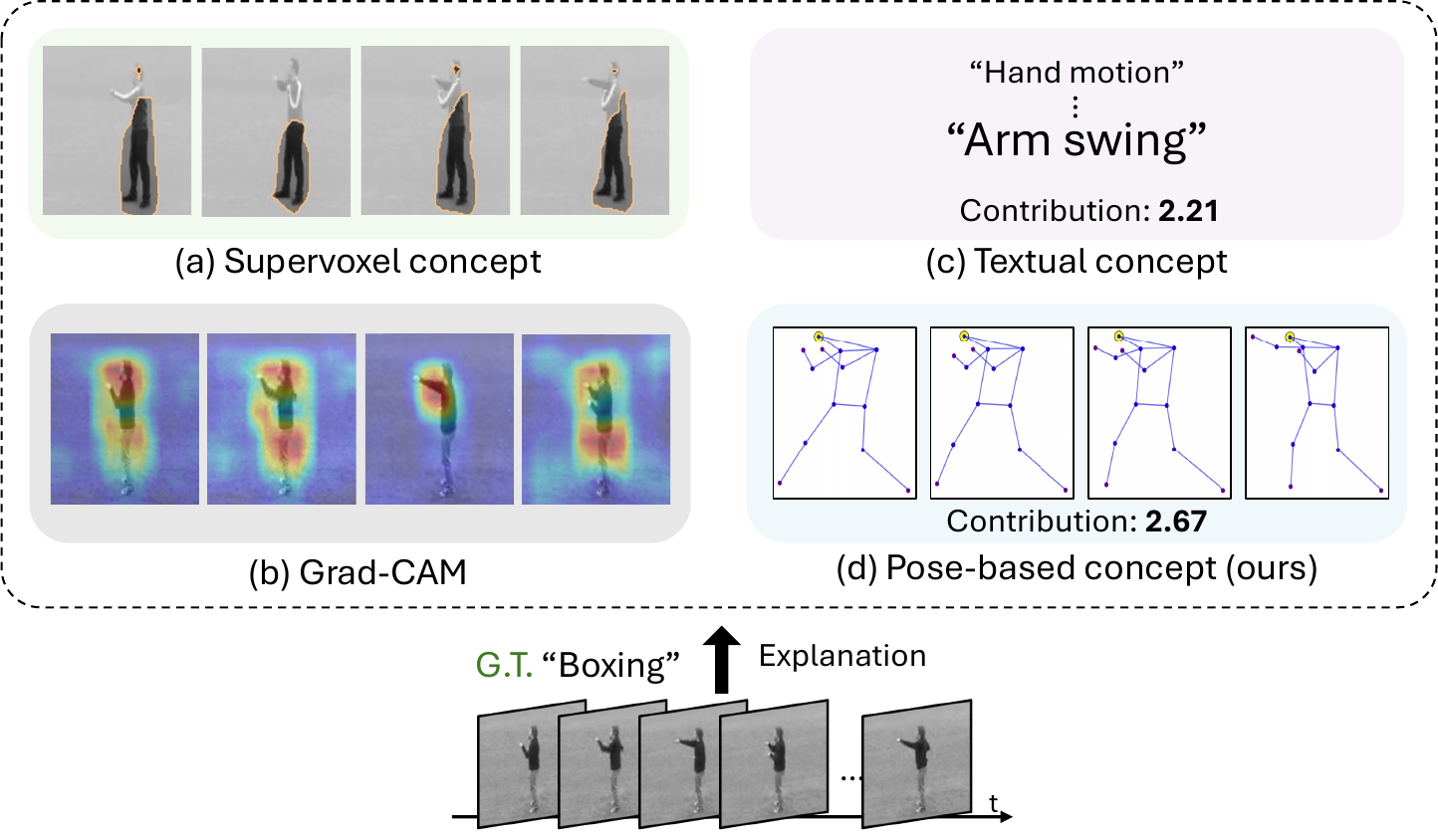}
    \captionsetup{justification=justified, singlelinecheck=false}
    \caption{\textbf{Comparison with existing XAI approaches for video action recognition.}  
    (a) Supervoxel-based concepts~\cite{vcad} cluster similar regions but lack motion dynamics.  
    (b) Grad-CAM~\cite{gradcam} highlights spatial regions but fails to capture temporal dependencies.  
    (c) Textual concepts~\cite{lf-cbm} provide abstract labels that miss fine-grained motion cues.  
    (d) \ours{} uses pose-based concepts to model spatial and temporal dynamics, offering interpretable, motion-aware explanations.}

    \label{fig:teaser}
\end{figure}
Advances in deep neural networks, including CNNs~\cite{resnet, googlenet, i3d, c3d, x3d, feichtenhofer2019slowfast} and transformers~\cite{vit, vivit, timesformer, motionformer, cast, videomae}, have significantly advanced human action recognition (HAR) performance.
However, these models remain opaque, making it difficult to understand their decision-making process.
Ensuring interpretability is crucial for reliability, accountability, and safety, particularly in real-world applications~\cite{doshi2017towards, jacovi2021formalizing}, leading to increased research in explainable AI (XAI) for HAR.

Existing video XAI methods primarily follow two paradigms: feature attribution and concept-based approaches.
Feature attribution methods~\cite{belharbi2023tcam, video-perturbation, occlusion_video, hartley2022swag, zeiler2014visualizing} localize spatiotemporal regions that influence model predictions.
However, they suffer from inherent limitations, such as sensitivity to perturbations and lack of robustness, often leading to unreliable explanations~\cite{craft, adebayo2018sanity, hase2021out, slack2020fooling}.

To overcome these issues, recent work explores Concept-based XAI (C-XAI), which explains models using high-level semantic concepts rather than pixel-level importance~\cite{vcad, 3d-ace, video-tcav}.
These methods define spatiotemporal concepts by clustering~\cite{slic} visually similar regions from videos.
However, as shown in \figref{teaser} (a), supervoxel-based concepts~\cite{vcad} highlight localized regions \eg lower body, but fail to capture underlying motion dynamics \eg stretching, making them ineffective for action recognition.
Additionally, these methods require per-sample optimization, making them computationally expensive and impractical for real-world applications~\cite{vcad, 3d-ace, video-tcav}.

An alternative is to adapt image-based XAI methods for HAR models.
One prominent category is feature attribution~\cite{lime, gradcam, perturbation, lrp, SHAP, fong2017interpretable, zeiler2014visualizing, cam}, which highlights spatially important regions that influence predictions.
However, as shown in \figref{teaser} (b), Grad-CAM~\cite{gradcam} predominantly highlights static human presence in each frame, failing to capture the temporal dependencies essential for action recognition.
To address this, Explainable-by-Design (EBD) methods~\cite{cbm, ProtoPNet, DCR, cem, lf-cbm, ta-cbm, pcbm, increcbm} enforce models to predict predefined semantic concepts \eg symbols, text, or prototypes before classification.
Unlike feature attribution, which assigns importance to low-level spatial features, EBD ensures structured interpretability by enforcing concept-based reasoning.
However, these methods primarily capture static appearance or object attributes rather than dynamic motion patterns, they struggle to capture fine-grained motion cues essential for action recognition.
For instance, applying text-based concepts to HAR often results in abstract explanations, such as \emph{Arm swing} as shown in \figref{teaser} (c), which fail to fully explain how the model predicts the action \emph{Boxing}.
These limitations highlight the necessity of explicitly motion-aware concept representations for explaining HAR models.

To address these challenges, we propose \textbf{P}ose \textbf{C}oncept \textbf{B}ottleneck for \textbf{E}xplainable \textbf{A}ction \textbf{R}ecognition (\ours{}), a novel concept bottleneck framework that enables transparent and interpretable human action recognition.
Unlike existing methods relying on feature attribution~\cite{gradcam, 3d-ace, video-tcav, cam, ace} or static textual concepts~\cite{lf-cbm, increcbm, ta-cbm, pcbm}, \ours{} provides structured explanations by explicitly modeling both spatial configurations and temporal motion dynamics.
To achieve this, \ours{} introduces two types of motion-aware pose-based concepts:
(i) \emph{static} pose-based concepts, which capture spatial configurations at individual frames, and
(ii) \emph{dynamic} pose-based concepts, which encode motion patterns across multiple frames.
Since human skeleton poses focus purely on body movements, they are inherently robust to static biases, such as background changes and lighting variations~\cite{stgcn, duan2022revisiting, shi2019skeleton, caetano2019skelemotion}.
Thanks to these properties, employing human pose as concepts enables \ours{} to effectively captures temporal dynamics, making it well-suited for providing interpretable, motion-aware explanations for action recognition.

Furthermore, to eliminate the need for manual annotations, we construct pose-based concepts using unsupervised clustering~\cite{finch}.
By grouping similar pose sequences automatically, \ours{} discovers meaningful concepts without explicit human labeling, ensuring robust concept discovery while avoiding the need for manual annotation.
This enables \ours{} to learn pose-based concepts effectively, making HAR transparent and interpretable without requiring skeleton inputs.

To validate the effectiveness of \ours{}, we conduct extensive experiments on three datasets where temporal context is critical for action recognition: KTH~\cite{kth}, Penn-Action~\cite{penn}, and HAA500~\cite{haa500}.
We quantitatively compare \ours{} to multiple baselines regarding concept efficiency and classification accuracy, demonstrating the strength of our method.
For interpretability evaluation, we qualitatively analyze pose-based concepts and their contributions at both the sample and group levels, clearly demonstrating how \ours{} reasons about actions.
Additionally, we validate our pose-based explanations through controlled experiments and demonstrate test-time interventions that modify model predictions in an interpretable manner.
The results demonstrate that \ours{} achieves strong action recognition performance while providing explicit, motion-aware explanations.

To summarize, our major contributions are as follows:
\begin{itemize}
\item We tackle the under-explored problem of explaining human action recognition. Instead of relying on RGB or textual concepts, we introduce human pose-based concepts as structured and motion-aware representations.
\item We propose \ours{}, a novel concept bottleneck framework that learns pose-based concepts directly from RGB videos without requiring explicit skeleton inputs, providing an interpretable representation of motion dynamics essential for action understanding.
\item We conduct extensive experiments to validate the effectiveness of \ours{}. Our results demonstrate that \ours{} provides effective pose-based explanations for human action recognition without performance degradation and enables test-time interventions for debugging model behavior.
\end{itemize}

\section{Related Work}
\label{sec:related}

\vspace{-0.5em}

\paragraph{Human action recognition.}
Human Action Recognition (HAR) is a core task in computer vision with applications in surveillance, healthcare, and human-computer interaction. Recent deep learning models, including CNNs~\cite{i3d, c3d, x3d, feichtenhofer2019slowfast} and Transformers~\cite{vivit, timesformer, motionformer, cast, videomae}, have advanced HAR by learning effective spatio-temporal representations.
However, despite their impressive accuracy, these models often function as black boxes, providing little insight into the rationale behind their predictions. This lack of transparency raises concerns in safety-critical applications, where interpretability is essential for ensuring trust and accountability~\cite{doshi2017towards, jacovi2021formalizing}. To address this, we propose \ours{}, a concept bottleneck framework that grounds model predictions in human-understandable, pose-based concepts, enabling more transparent and interpretable HAR.
\vspace{-1em}

\paragraph{Explainable AI in video domain.}
While HAR has seen significant progress, explainability in the video domain remains under-explored compared to the image domain.
Early video XAI efforts adapted feature attribution methods from image XAI approaches~\cite{belharbi2023tcam, video-perturbation, occlusion_video, hartley2022swag}.
For instance, TCAM~\cite{belharbi2023tcam} extends class activation mapping (CAM) by incorporating temporal pooling to highlight important regions across frames.
However, these methods fail to provide complete explanations and often require additional interpretation to extract meaningful insights.
To address this, recent works in Concept-based XAI (C-XAI)\cite{vcad, 3d-ace, video-tcav} have explored high-level semantic explanations using post-hoc concept discovery.
For example, 3D-ACE\cite{ace} applies supervoxel-based clustering to discover spatio-temporal concepts, while Video-TCAV~\cite{video-tcav} extends TCAV~\cite{tcav} to the video domain by identifying concept importance across temporal sequences.
Although these methods cluster spatiotemporal segments, they are often limited by their reliance on static visual features, missing fine-grained motion patterns critical for action recognition.
Moreover, as post-hoc methods, they require computationally expensive optimizations, limiting scalability.
In contrast, \ours{} leverages a concept bottleneck framework~\cite{cbm}, reasoning through pose-based concepts derived from skeleton sequences, capturing motion patterns effectively and without the need for additional optimization.
\vspace{-1.5em}

\paragraph{Concept-based explanation.}
Concept-based explanations enhance interpretability by linking model predictions to human-understandable concepts~\cite{achtibat2023attribution}.
These methods can be broadly categorized into \emph{post-hoc} concept discovery~\cite{ahn2024unified, tcav, ibd, ace, craft}, which extracts concepts after model training, and \emph{explainable-by-design} approaches~\cite{cbm, cem, DCR, ProtoPNet}, which integrate concept layers directly into the model structure.
CBM~\cite{cbm} enforces interpretability by introducing a concept layer that predicts predefined human-interpretable concepts before final classification.
Recent works~\cite{lf-cbm, increcbm, pcbm} further leverage Vision-Language Models (VLMs) to obtain textual concepts without manual labeling.
However, most of these methods focus on image domains, which struggle to capture the temporal dependencies crucial for human action recognition.
To address this limitation, \ours{} introduces human skeleton poses as symbolic concepts, enabling structured, motion-aware explanations tailored for human action recognition.

\setlength{\abovedisplayskip}{5pt}
\setlength{\belowdisplayskip}{5pt}

\section{\ours{}}
\label{method}

\begin{figure*}[t]
    \centering
    \includegraphics[width=\linewidth]{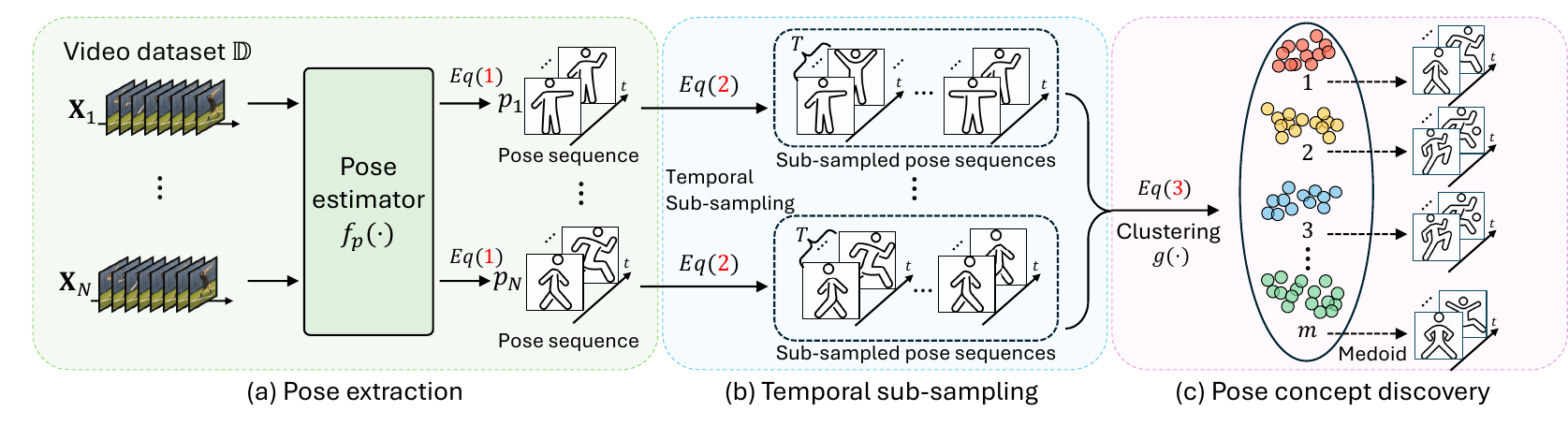}
    \captionsetup{justification=justified, singlelinecheck=false}
    \caption{\textbf{Overview of pose-based concept construction.}  
    (a) We extract a pose sequence from each video using the pose estimator $f_p(\cdot)$. 
    (b) Since video frames are too long to be directly used as concepts, we perform temporal sub-sampling to obtain overlapping sub-sampled pose sequences of length $T$.  
    (c) We apply clustering to all pose sub-sampled pose sequences in the dataset to discover pose-based concepts.  
    (d) The medoid of each cluster serves as the representative pose, ensuring interpretability by selecting actual samples from the dataset.}
    \label{fig:make_concept}
\end{figure*}

We propose the Pose Concept Bottleneck Model (\ours{}), a novel explainable framework for interpretable human action recognition that leverages pose-based human concepts.
Human skeleton poses represent a time series of joint locations in 2D or 3D space, capturing the structural configuration and movement patterns of the human body~\cite{stgcn,shi2019skeleton,caetano2019skelemotion}.
Unlike raw RGB features, which encode both appearance and background context, skeleton-based representations focus solely on motion dynamics, making them robust to background variations, lighting conditions, and occlusions~\cite{duan2022revisiting}.

By focusing on pose-based representations, we ensure that the learned concepts align closely with human perception of actions, capturing both spatial structure and temporal dynamics.
Since manually annotating pose-based concepts is impractical for large-scale datasets, we introduce a simple yet efficient approach to automatically discover these concepts from RGB videos.
Specifically, our framework involves:
(i) extracting skeleton poses from RGB videos at the clip level,
(ii) clustering these poses into pose-based concepts, and
(iii) training a concept bottleneck model.
Thanks to this data-driven approach, \ours{} provides inherently interpretable, motion-aware explanations aligned closely with human perception.

In the following sections, we formalize the problem setting (\secref{formulation}), describe the process of pose-based concept construction (\secref{annotation}), and present the CBM training strategy (\secref{training}).

\subsection{Problem Formulation}
\vspace{-0.5em}

\label{sec:formulation}
The goal of \ours{} is to learn a concept layer $f_c(\cdot;W_C)$ and a classifier $\phi(\cdot;W_F)$ on top of a pre-trained and frozen feature extractor $f_v(\cdot)$.
Formally, we define a dataset $\mathbb{D} = \{(X_i, c_i, y_i)\}_{i=1}^{N}$, where $X_i$ is the $i$-th video, $c_i$ is the corresponding concept label, and $y_i$ is the action label.
Formally, given a dataset $\mathbb{D}=\{(X_i,c_i,y_i)\}_{i=1}^{N}$ consisting of $N$ video samples, we uniformly sample $l$ frames from each video $X_i \in \mathbb{R}^{L\times C\times H\times W}$ to form an input clip $x_i \in \mathbb{R}^{l\times C\times H\times W}$.
Here, $L$ denotes the number of video frames, while $l$, $C$, $H$, and $W$ represent the number of sampled frames, RGB channels, height, and width, respectively.
Each video $X_i$ is associated with a multi-label pose-based concept vector $c_i \in \{0,1\}^{m}$ and an action label $y_i \in \mathbb{R}^{k}$, where $m$ denotes the number of pose-based concepts and $k$ denotes the number of action classes.

\subsection{Pose-Based Concept Construction}
\vspace{-0.5em}

\label{sec:annotation}
To provide a structured representation of motion, we use human poses as our concept for interpreting human action recognition (HAR)~\cite{duan2022revisiting,shi2019skeleton,caetano2019skelemotion,du2015hierarchical}.
Unlike RGB frames, human skeleton poses focus on movement while remaining robust to appearance changes and background variations.
To capture both spatial configurations and temporal dynamics, \ours{} defines two types of pose-based concepts:
(i) Static pose-based concepts, representing a frame-level pose capturing static pose, and
(ii) Dynamic pose-based concepts, representing pose sequence capturing motion patterns over multiple frames.
To assign these concepts to each video, we extract skeleton pose sequence and group them into meaningful clusters, allowing \ours{} to learn pose-based representations without manual labeling.

\vspace{-1.5em}
\paragraph{Pose extraction.}
We use an off-the-shelf pose estimator~\cite{pose_estimator} $f_p(\cdot)$ to extract skeleton-based representations from videos.  
As illustrated in \figref{make_concept} (a), given a dataset $\mathbb{D}$ consisting of $N$ video samples, we pass each video $X_i \in \mathbb{R}^{L\times C\times H\times W}$ through $f_p(\cdot)$ to obtain pose sequences $\mathbf{P} = \{p_i\}_{i=1}^{N}$ as follows:
\vspace{-0.7em}
\begin{equation}
p_i = f_p(X_i),
\end{equation}
where $p_i \in \mathbb{R}^{L \times J \times 2}$ represents the extracted 2D pose sequence of $X_i$, with $L$ denoting the number of video frames, $J$ the number of joints. 
\begin{figure*}[t]
    \centering
    \includegraphics[width=0.8\linewidth]{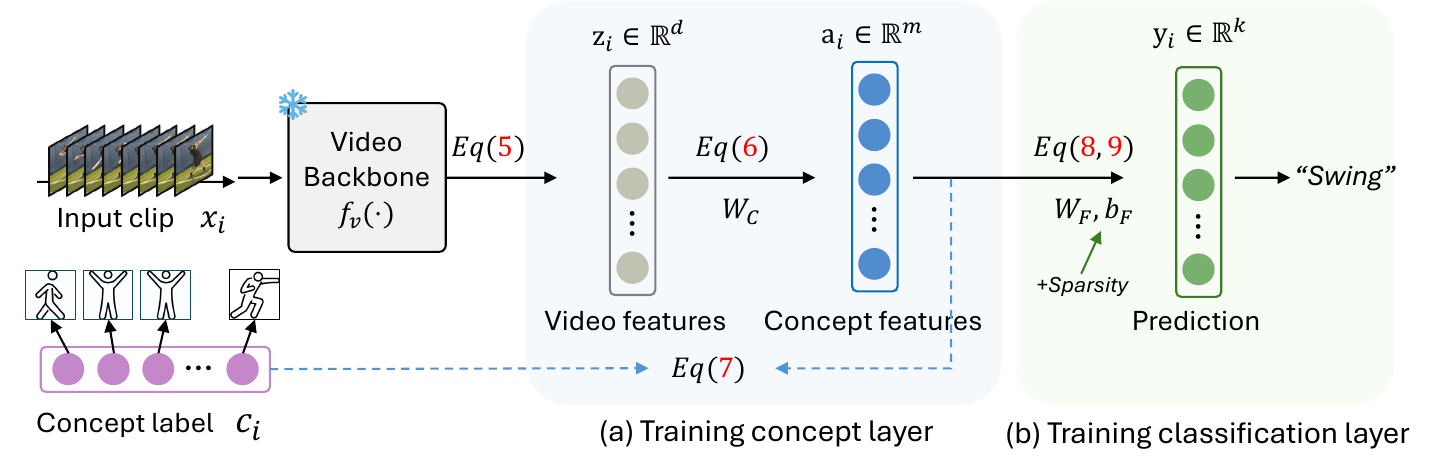}
    \captionsetup{justification=justified, singlelinecheck=false}
    \caption{\textbf{Training overview of \ours{}.}  
    (a) Given an input clip $x_i$, we extract video features $z_i$ using the video backbone $f_v(\cdot)$, then map them to concept features $a_i$ via the learnable concept projection matrix $W_c$. The model optimizes $W_c$ using the cosine cubed loss to align $a_i$ with the assigned concept label $c_i$.  
    (b) Given concept features $a_i$, the classifier parameterized by $W_F$ and $b_F$ predicts $\hat{y}_i$. To improve interpretability, we employ a sparsity regularization~\cite{sparselinear} in the classification layer.}
    \label{fig:training}
\end{figure*}
\vspace{-1.5em}

\paragraph{Temporal sub-sampling.}  
Since the number of video frames \(L\) is too long to be used directly as a concept, we perform temporal sub-sampling on each pose sequence, as shown in \figref{make_concept} (b). 
Given a pose sequence \(p_i \in \mathbb{R}^{L \times J \times 2}\), we sample overlapping pose sequences of length \(T\) along the temporal axis, where \(T\) is a predefined window size and \(L > T\).  
We define the \(t\)-th sub-sampled pose sequence as: 
\begin{equation}
p_{i}^{(t)} = \{ p_{i,t}, p_{i,t+1}, \dots, p_{i,t+T-1} \},
\end{equation}
where $p_{i}^{(t)} \in \mathbb{R}^{T \times J \times 2}$ represents a sub-sampled pose sequence from $p_i$.
Here, when \(T = 1\), a sub-sampled pose represents a \emph{static} pose-based concept, capturing spatial relationships among the joints.  
For \(T > 1\), it forms a \emph{dynamic} pose-based concept, encoding motion patterns over multiple frames.  
Both types of pose-based concepts are crucial for action recognition~\cite{stgcn}, as human actions involve both distinct postures and continuous motion dynamics.  
\vspace{-1.5em}
\paragraph{Pose-based concept discovery via clustering.}  
As illustrated in \figref{make_concept} (c), we apply clustering~\cite{finch} to all sub-sampled pose sequences in the dataset, grouping them into $m$ clusters to define pose-based concepts. 
Formally, we define a cluster assignment function $g(\cdot)$ that maps each sub-sampled pose sequence $p_{i}^{(t)}$ to one of the $m$ clusters:
\begin{equation}
    g(p_{i}^{(t)}) \in \{1, 2, \dots, m\}.
\end{equation}
To obtain the concept label for video $X_i$, we aggregate the assigned cluster indices from all its sub-sampled pose sequence into a set:
\begin{equation}
    \Omega_i = \bigcup_{t} g(p_{i}^{(t)}),
\end{equation}
where $\Omega_i$ contains the unique concept indices present in $X_i$.  
Finally, we obtain the multi-label pose-based concept vector $c_i \in \mathbb{R}^{m}$, setting each element to $1$ if its corresponding cluster index appears in $\Omega_i$.
\vspace{-1.5em}
\paragraph{Selecting representative pose-based concept.}  
We obtain $m$ pose-based concepts through clustering, allowing us to annotate videos with concept labels.  
However, clustering alone does not specify what each concept represents in terms of actual human poses.  
To address this, we define representative poses and pose sequences for static and dynamic concepts, respectively, by selecting their medoids the most central samples within the cluster.
Please refer to the \figref{concept_visualization} for visualizations of these representative poses.

\subsection{Training} 

\label{sec:training}
Once concept annotations are obtained, we train \ours{} to predict human pose-based concepts and classify actions.  
\ours{} comprises two key components:  
(i) \emph{concept layer} that maps video features to pose-based concepts, and  
(ii) \emph{classification layer} that predicts actions based on the learned concepts.
\paragraph{Training concept layer.}  
\vspace{-1.5em}
As illustrated in \figref{training} (a), given an input clip $x_i$ and its corresponding concept label $c_i$, we first extract video features by passing $x_i$ through a video backbone $f_v(\cdot)$, producing $z_i \in \mathbb{R}^{d}$:  
\begin{equation}
    z_i = f_v(x_i).
\end{equation}
We then project $z_i$ onto the concept space using the learnable concept projection matrix $W_c \in \mathbb{R}^{d \times m}$:
\begin{equation}
    a_i = f_c(z_i;W_C) = W_c^\top z_i.
\end{equation}
We standardize both $a_i$ and $c_i$ to obtain $\bar{a}_i$ and $\bar{c}_i$, ensuring zero mean and unit variance.  
Then, we optimize $W_c$ using the \emph{cosine cubed} loss~\cite{Dissect} as follows:
\begin{equation}
    \min_{W_c} \sum_{i \in \mathbf{B}} L_c(\bar{a}_i, \bar{c}_i),
\end{equation}
where $\mathbf{B}$ represents a mini-batch of training samples.
\vspace{-1.em}
\paragraph{Training classification layer.}  
As illustrated in \figref{training} (c), after training the concept layer, we freeze the concept projection matrix $W_C$ and train only the classification layer.  
The classification layer is parameterized by a learnable weight matrix $W_F \in \mathbb{R}^{m \times k}$ and a bias $b_F$.  
To enhance interpretability, we employ a sparse linear layer~\cite{sparselinear}.  
Given normalized concept features $\bar{a}_i$, the classifier $\phi(;W_F,b_F)$ predicts the action label $\hat{y}_i$:  
\begin{equation}
    \hat{y}_i = \phi(\bar{a}_i;W_F,b_F) = W_F^\top \bar{a}_i + b_F.
\label{eq:prediction}
\end{equation}  
We optimize $W_F$ by minimizing the cross-entropy loss $L_{\text{ce}}$ with an additional sparsity regularization term $R(W_F)$:  
\begin{equation}
    \min_{W_F, {b_F}} \sum_{i\in\mathbb{B}} L_{\text{ce}} (\hat{y}_i, y_i) + \lambda R(W_F).
\end{equation}
By training both the concept layer and the classifier with sparsity regularization, \ours{} learns to recognize actions through pose-based concepts, ensuring that predictions are grounded in interpretable motion patterns rather than raw visual features.
This structured reasoning enables more transparent and explainable human action recognition.
\section{Experimental Results}
\label{sec:results}

In this section, we first introduce the datasets and implementation details in (\secref{setup}). We then present quantitative results in (\secref{quantity}), evaluating accuracy, efficiency, and clustering performance using the NMI score. In (\secref{quality}), we provide qualitative analyses, including pose-based concept visualizations and weight interpretation. Finally, in (\secref{discission}), we discuss the impact of interventions, group-level concept analysis, and whether the model truly learns pose representations.

\subsection{Experimental setup}
\label{sec:setup}
\begin{table*}[t]
    \scriptsize
    \captionsetup{justification=justified, singlelinecheck=false}
    \centering
    \caption{\textbf{Evaluation of baseline methods with varying concept types and numbers across multiple datasets.} We report the number of concepts , concept utilization efficiency and Top-1 accuracy (\%). The best are in \best{bold}. }

    \vspace{-1em}    
    \resizebox{0.8\linewidth}{!}{
    \begin{tabular}{l l rrr rrr rrr  }
        \toprule

        \multirow{4}{*}{Method} & 
        \multirow{4}{*}{Concept type.} &
        \multicolumn{3}{c}{KTH-5~\cite{kth}}& 
        \multicolumn{3}{c}{Penn action~\cite{penn}} & 
        \multicolumn{3}{c}{HAA-49~\cite{haa500}} \\

    \cmidrule(lr){3-5}
    \cmidrule(lr){6-8}
    \cmidrule(lr){9-11}

                                        && NUM. & CUE &Acc. & NUM. & CUE &Acc.& NUM. & CUE &Acc. \\
        
        \midrule
    \multirow{2}{*}{VideoMAE(B/16)~\cite{videomae}}               &None  & -  & - & 96.52 & - & - & 95.31 & - & -& 79.57 \\
                                               &Textual & 30 & 319.9 & 95.96 & 80 &121.9 & 97.49 & 300 & 25.9 & 77.59 \\ 
                                           \midrule
    \multirow{4}{*}{\ours{}(B/16)}&Static     & 30  & 317.6 & 95.27 & 80 & \best{122.2} & 97.74 & 300 & \best{27.0} & 81.12 \\
                                                             &Static     & 3324  & 2.9 & \best{96.94} & 170 & 57.6 & 97.84 & 2046 & 4.0 & 82.65\\
                                                             &Dynamic     & 30  & \best{320.3} & 96.10 & 80 & 121.6 & 97.30 & 300 & 26.2 & 78.73 \\
                                                            &Dynamic     & 257  & 37.6 & 96.51 & 328 & 29.8 & \best{97.90} & 2145 & 3.9 & \best{83.90}\\

        \bottomrule
    \end{tabular}
    }
    \label{tab:acc_cue}
    \vspace{-1.5em}
\end{table*}

\paragraph{Datasets.}
We evaluate \ours{} on three video datasets where temporal context is crucial: KTH-5, Penn Action, and HAA49.
i) KTH-5 is a refined version of the KTH dataset~\cite{kth}, consisting of five distinct action classes after removing the “jogging” class due to overlap with “walking” and “running.”
ii) Penn Action~\cite{penn} contains 2,326 videos across 15 action classes with detailed human joint annotations, ideal for both action recognition and pose estimation.
iii) HAA49 is a subset of the HAA500 dataset~\cite{haa500}, where we manually select 49 human-centric action classes that rely heavily on temporal context.
Please refer to the \appenref{dataset}.
\vspace{-1.5em}
\paragraph{Implementation details.}
We implement our framework using PyTorch, built upon the Label-Free CBM~\cite{lf-cbm} code base.
We use VideoMAE-B/16~\cite{videomae} as our frozen feature extractor, pretrained on Kinetics-400~\cite{kinetics}.
We use the off-the-shelf pose estimator ViTPose-B~\cite{vitpose} pretrained on COCO~\cite{coco}, leveraging 17 COCO keypoints for pose extraction.
We create the concept set for the textual concept baseline in \figref{teaser} and \tabref{acc_cue}, following the Label-Free-CBM method~\cite{lf-cbm}.
Please refer to the \appenref{implementations}.
\vspace{-1.5em}
\paragraph{Evaluation metric.}
To assess the efficiency of concept utilization in \ours{}, we adapt the Concept Utilization Efficiency (CUE) metric from Res-CBM~\cite{increcbm}, with the concept length $\bar{L}$ fixed to $1$.
Additionally, we use the Normalized Mutual Information (NMI) score to evaluate the alignment between learned concept clusters and ground truth action labels, ensuring meaningful concept discovery.
Please refer to the \appenref{metric}.

\subsection{Quantitative Results}
\label{sec:quantity}
\paragraph{Accuracy and concept efficiency. }

We evaluate the Top-1 accuracy of \ours{} across three datasets using VideoMAE(B/16)~\cite{videomae} as the backbone, as shown in \tabref{acc_cue}. The baseline is the original VideoMAE model. We compare the performance of different concept types: dynamic pose, static pose, and text.
\ours{} consistently outperforms the baseline across all datasets, demonstrating improved performance. Notably, \ours{} achieves high CUE with fewer concepts, particularly on KTH-5 and Penn Action datasets. Dynamic pose-based concepts perform better than static ones, offering superior action differentiation, especially for visually similar movements. Text-based concepts underperform, as they fail to capture the necessary motion details for action recognition.
Given that dynamic pose-based concepts preserve temporal structure and enhance action differentiation, we use them in subsequent experiments.
\begin{table}[t]
    \centering
    \captionsetup{justification=justified, singlelinecheck=false}
    \caption{\textbf{Evaluation of baseline methods with varying the number of concepts on the Penn Action dataset.} We report the number of concepts, concept utilization efficiency, and Top-1 accuracy (\%). The number of concepts in each partition is shown in parentheses. The best are in \best{bold}.}
 \resizebox{0.8\linewidth}{!}{
    
    \begin{tabular}{lccc}
        \toprule
        Partition (Concepts) & NMI Score & Acc. & CUE \\
        \midrule
        Partition 0 (1621)  & 49.07  & \best{97.90}  &  6.0 \\
        Partition 1 (328)   & \best{50.78}  & \best{97.90}  &  29.9 \\
        Partition 2 (63)   & 47.90 & 96.70  &  153.5 \\
        Partition 3 (17)    & 42.10  & 93.50  &  \best{550.0} \\
        Partition 4 (3)    & 16.67  & N/A  &N/A  \\
        \bottomrule
    \end{tabular}
    }
    \label{tab:nmi}
\end{table}

\vspace{-1em}
\paragraph{Clustering NMI score. }
We evaluate pose clustering on the Penn Action dataset using the FINCH algorithm~\cite{finch} with VideoMAE(B/16)~\cite{videomae}. 
The FINCH algorithm hierarchically clusters the dataset into multiple partitions, allowing us to control the number of clusters and analyze the impact of different concept groupings on action recognition performance.
As shown in \tabref{nmi}, increasing the number of concepts improves accuracy but lowers Concept Utilization Efficiency (CUE).
When the number of concepts is too small, particularly fewer than the number of classes, the model fails to distinguish actions.
For a best trade-off between accuracy and efficiency, we can select Partition 2 with 63 concepts, which provides the best performance for the Penn Action dataset with 15 classes.

\begin{figure}[t]
\includegraphics[width=\linewidth]{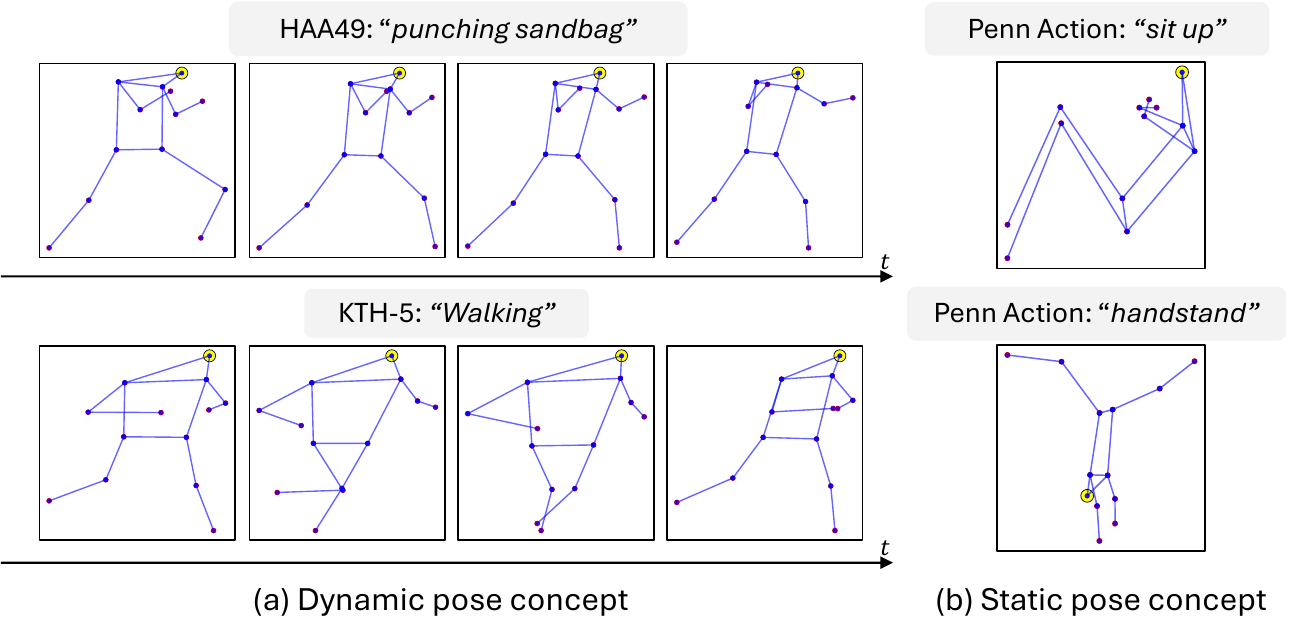}
\centering
    \captionsetup{justification=justified, singlelinecheck=false}
    \caption{\textbf{Visualizations of representative pose-based concept.} (a) Dynamic pose concepts represent motion patterns across multiple frames. (b) Static pose concepts represent spatial configurations  We also show the corresponding video class for each concept.}
\label{fig:concept_visualization}
\end{figure}

\subsection{Qualitative Results}
\label{sec:quality}
\paragraph{Pose-based concept visualization. }

To visualize pose-based concepts, we define a representative pose for each concept by selecting its medoid in \secref{annotation}. When visualizing the concepts, we use the representative pose sequence or pose. In \figref{concept_visualization} (a), we visualize dynamic pose-based concepts for the “punching sandbag” class from HAA49~\cite{haa500} and the “walking” class from KTH-5~\cite{kth}.
We capture the dynamic movement of the fist and the act of walking, showing that dynamic pose-based concepts effectively represent sequential actions. In \figref{concept_visualization} (b), we show static pose-based concepts for the “sit up” class from Penn Action and the “handstand” class from HAA49, clearly depicting key moments of these actions. 
These results demonstrate that \ours{} captures meaningful motion representations, improving interpretability and distinguishing actions.

\vspace{-1em}
\begin{figure}[t]
\includegraphics[width=\linewidth]{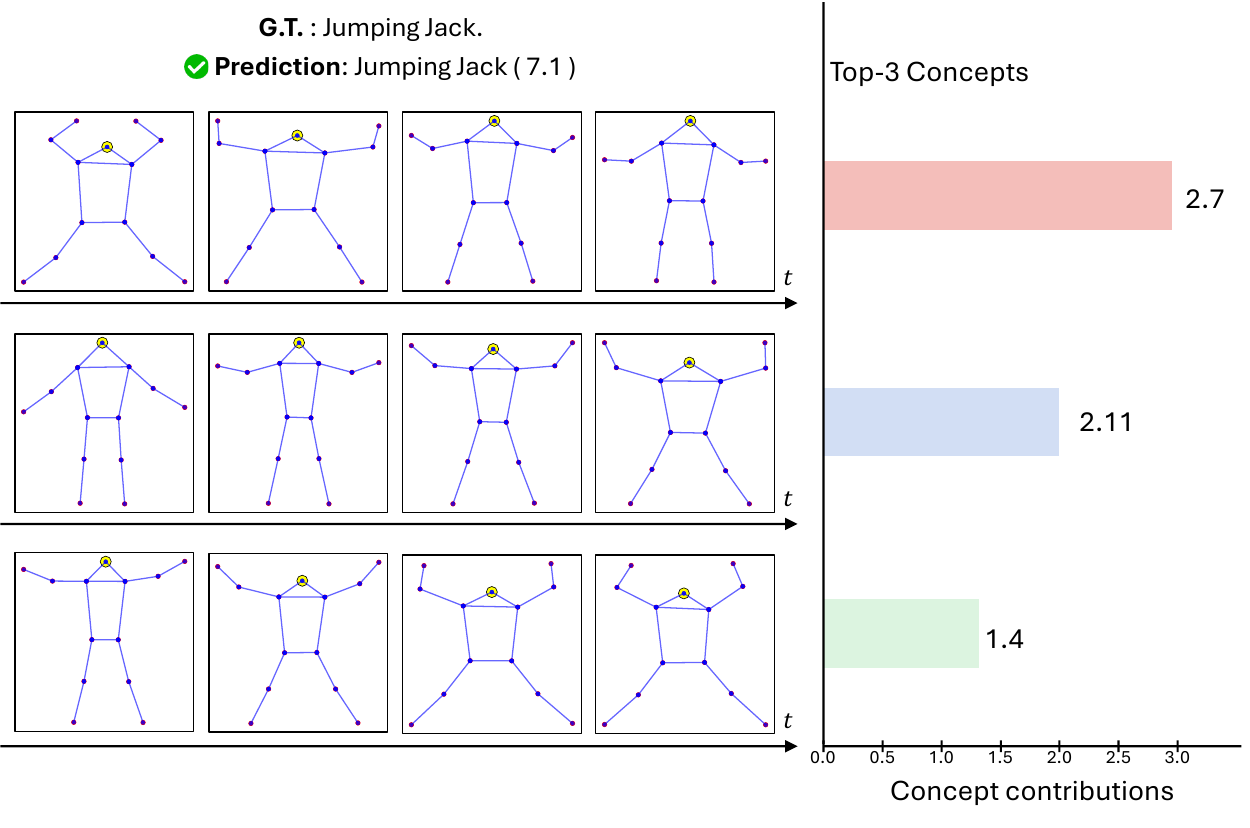}
\centering
    \captionsetup{justification=justified, singlelinecheck=false}
    \caption{\textbf{Concept contributions.} We visualize the dynamic pose concepts corresponding to the top-3 concept contributions for a sample from ``Jumping Jack'' class.}
\label{fig:concept_contribution}
\end{figure}
\paragraph{Concept-contributions.  }
In \ours{}, we compute the concept contributions following the approach in \cite{lf-cbm}. The contribution of concept $i$ to output class $j$ on input $x_k$ is computed as $\text{Contrib}(x_k, i, j) = W_F[i, j] \cdot fc(x_k)$, where $W_F \in \mathbb{R}^{m \times k}$ is the weight matrix from \eqnref{prediction}. Since $W_F$ is sparse, most of the contributions are zero, making it easy to visualize the important ones. In \figref{concept_contribution}, we visualize the concept contributions for a sample from the “Jumping Jack” class in the Penn Action dataset~\cite{penn}. The top-contributing pose-based concepts effectively capture the repetitive motion of extending arms and legs, which aligns with the Jumping Jack movement. This demonstrates how \ours{} successfully learns and prioritizes key motion patterns for action recognition.
\vspace{-1em}
\begin{figure}[t]
    \centering
    \captionsetup{justification=justified, singlelinecheck=false}
    \includegraphics[width=\linewidth]{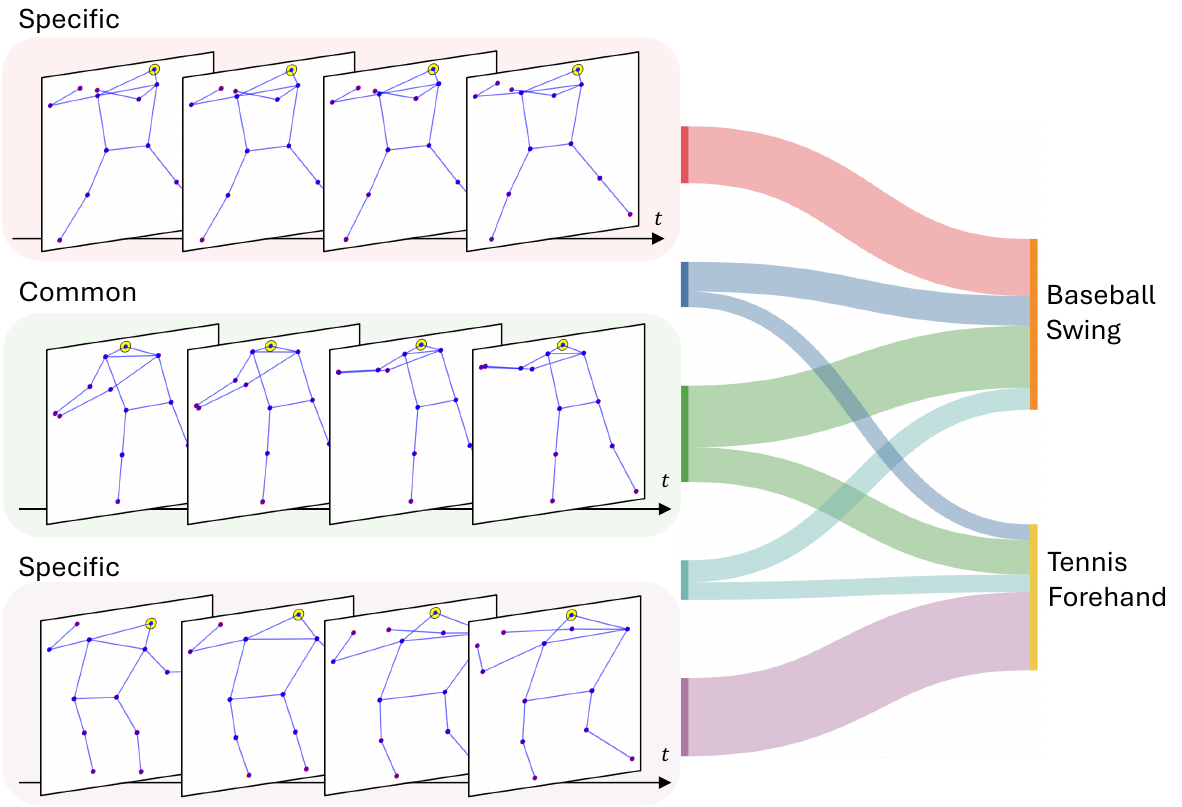}
    \caption{\textbf{Weight visualization.} We show the final layer weight distribution between concepts and action classes, focusing on ``Baseball Pitch'' and ``Tennis Serve''. To enhance clarity in the figure, we show only the \emph{common} concept with the largest weight. }    
\label{fig:weight_visualize}
\end{figure}

\paragraph{Weight visualization. }

In \figref{weight_visualize}, we visualize the weights connecting pose-based concepts to action classes in \ours{} using the Penn Action dataset~\cite{penn}.
We show the learned weights for the ``Baseball Swing'' and ``Tennis Forehand'' classes, highlighting how both \emph{common} and \emph{specific} concepts contribute to these actions.
Both specific concepts capture important pose sequences for each action. 
For instance, the specific concept for ``Tennis Forehand'' captures a pose sequence of swinging with one hand.
Interestingly, the common pose-based concept captures a shared pose sequence important for both actions, reflecting the fundamental swinging motion common to both.
These visualizations offer insights into how \ours{} learns to represent action recognition by capturing the motion patterns relevant to each action. 
Notably, the shared knowledge between actions aligns with human intuition, highlighting the model’s ability to generalize between similar actions.

\begin{figure}[t]
    \centering
    \captionsetup{justification=justified, singlelinecheck=false}
    \includegraphics[width=\linewidth]{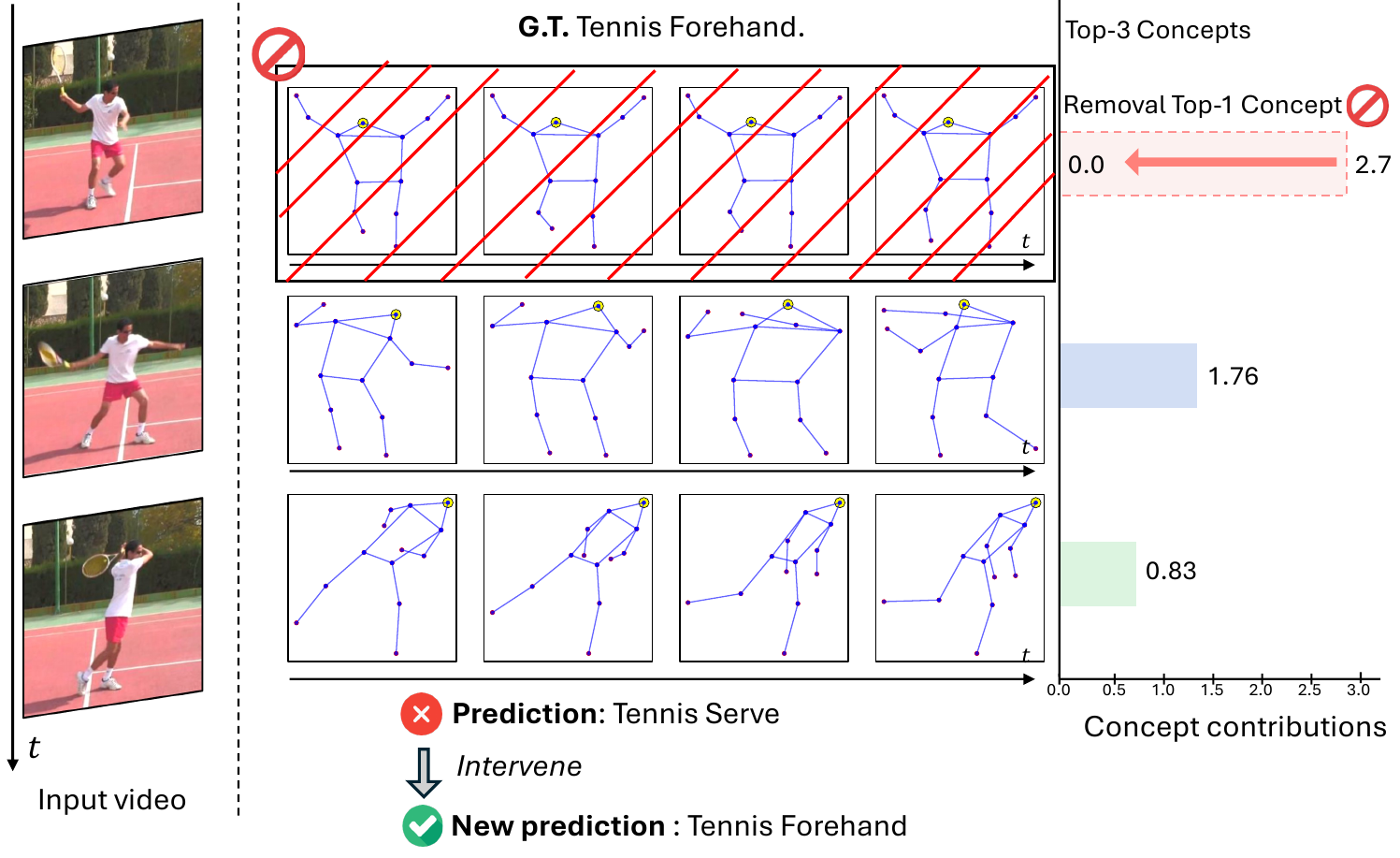}
    \caption{\textbf{Test-Time Intervention.} Given an input video of a ``Tennis Forehand'' action, the model initially fails to predict the correct action, highlighting an irrelevant top-1 concept contribution. By removing the top-1 concept, we eliminate the incorrect concept and enable the model to correctly predict the ``Tennis Forehand'' action based on the top-2 relevant concepts.} 
\label{fig:intervention_new}
\end{figure}

\subsection{Discussion}
In this section, we conduct experiments using the Penn Action~\cite{penn} dataset with VideoMAE-B/16~\cite{videomae}.
\label{sec:discission}
\vspace{-1.5em}

\paragraph{Intervention.} 

In \figref{intervention_new}, we present an example where the model initially misclassifies a “Tennis Forehand” action as a “Tennis Serve”. 
We suppress the activation of the concept with the highest contribution to the model’s decision, which was leading the model to misclassify the “Tennis Forehand” action as “Tennis Serve".
This pose-based concept is unrelated to the Tennis Forehand action, so we set its activation to zero to correct the prediction.
Remarkably, this intervention causes the model to change its prediction, demonstrating that even modifying the activation of a single, highly-contributing concept can significantly influence the final classification.
This finding underscores the effectiveness of concept-based interventions in refining model predictions. 
By directly controlling concept activations, we gain deeper insights into the model’s decision-making process. 
This targeted adjustment not only enhances interpretability but also reduces confusion between similar actions, showing how concept-based reasoning can improve action recognition accuracy.

\begin{figure}[t]
\includegraphics[width=\linewidth]{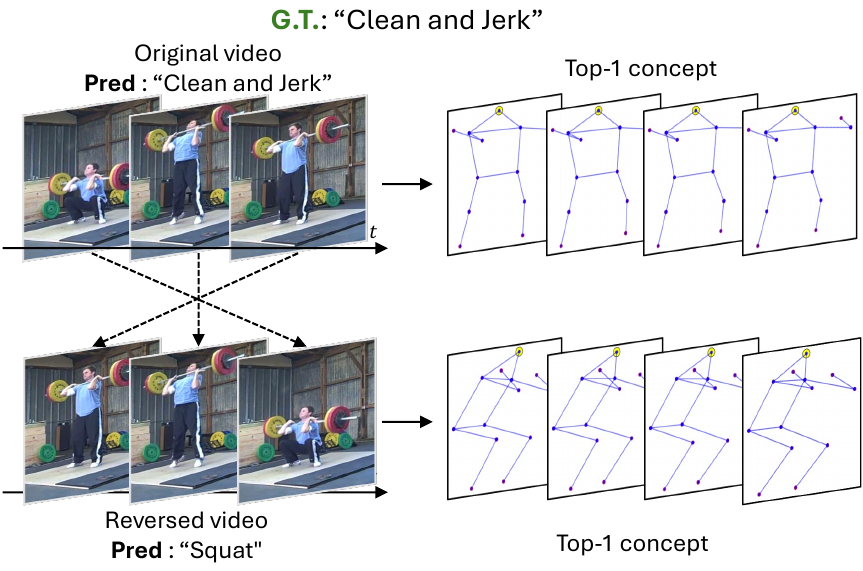}
    \captionsetup{justification=justified, singlelinecheck=false}
    \caption{\textbf{Evaluation of temporal learning using reversed sequences. } The figure illustrates an experiment where the temporal order of test videos is reversed to assess whether \ours{} effectively captures temporal dependencies.}
\label{fig:reverse}
\end{figure}
\begin{figure}[t]
    \includegraphics[width=0.9\linewidth]{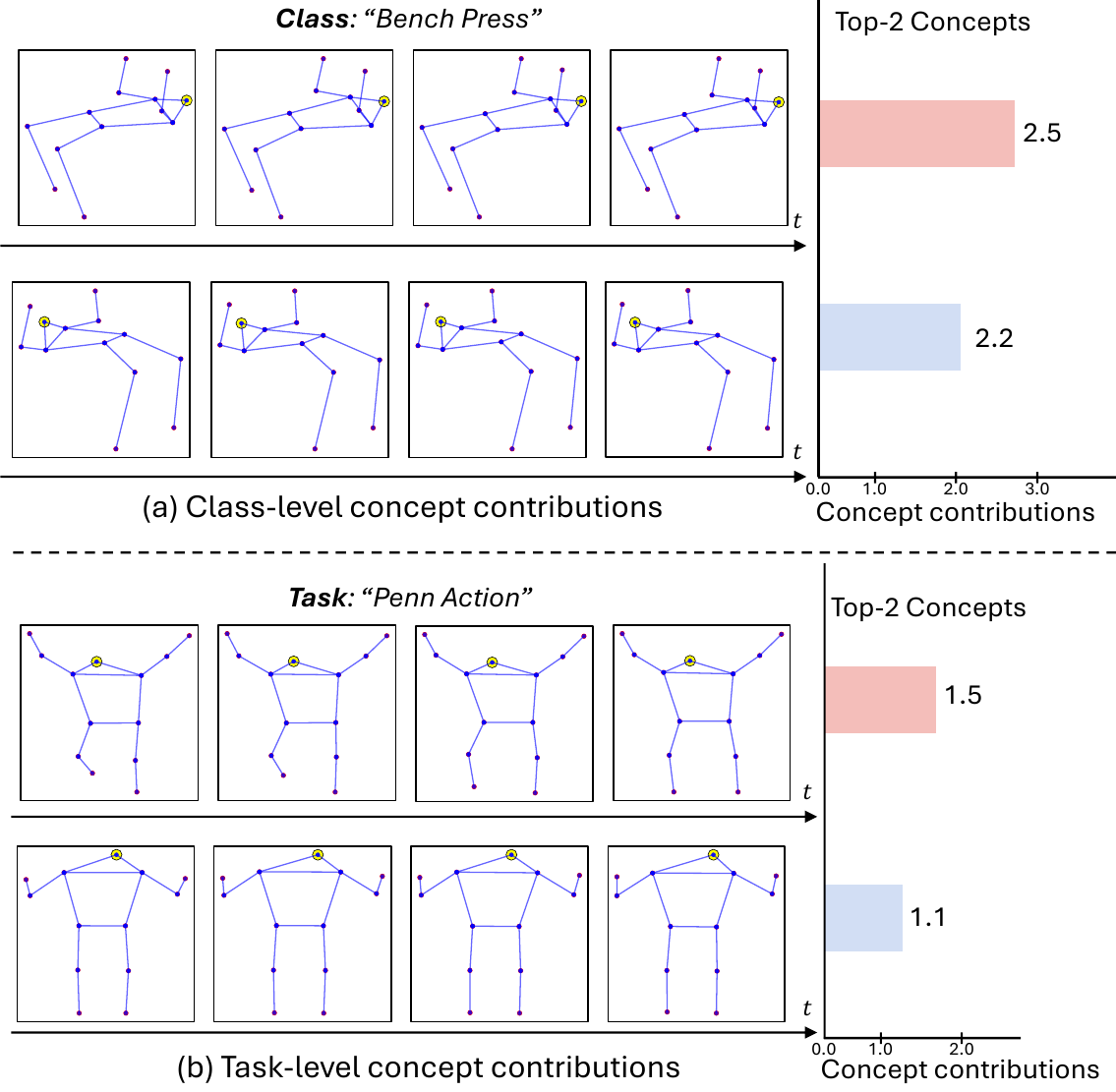}
    \centering
    \captionsetup{justification=justified, singlelinecheck=false}
    \caption{\textbf{Visualizations of class-level and task-level concept contributions.} (a) Class-level contributions for “Bench Press” and (b) task-level contributions for “Penn Action” dataset with the top-2 concepts and their contributions. }
    \vspace{-1em}
\label{fig:group_level}
\end{figure}

\vspace{-1em}
\paragraph{Temporal context in pose-based concepts. }
To evaluate how well \ours{} captures temporal context, we an conduct experiment using reversed video sequences as a test input. 
As shown in \figref{reverse}, the original video depicts a ``Clean and Jerk'' action, but when played in reverse, the model classifies it as ``Squat''.
This misclassification occurs because the ``Clean and Jerk'' motion involves lifting a barbell while standing up, whereas the reversed sequence resembles lowering a barbell into a squatting position.
The fact that the model predicts ``Squat'' for a reversed ``Clean and Jerk'' sequence demonstrates that it captures the sequential nature of pose-based concepts rather than relying solely on static poses.
This finding highlights that \ours{} effectively captures temporal structures, utilizing motion order to differentiate between actions and reinforcing its ability to model temporal context in human motion understanding.
\vspace{-1em}

\paragraph{Group level analysis. }

In \figref{group_level}, we visualizes class-level concept contributions for Bench Press and task-level concept contributions for Penn Action~\cite{penn}. 
While sample-level concept contribution analysis provides insights into individual predictions, it does not reveal consistently important concepts across samples. 
To address this, we conduct class-level and task-level concept contribution analyses. 
Class-level analysis aggregates and normalizes concept contributions from all samples of a given class to identify key concepts crucial for distinguishing actions.
In \figref{group_level} (a), we see that the model assigns high contributions to concepts related to a lying posture with vertical arm movements, aligning with the bench press motion.
Notably, the top three concepts with the highest contributions are all strongly associated with the bench press action, emphasizing that \ours{} effectively captures specific motion features for class recognition.
In \figref{group_level} (b), task-level analysis sums and normalizes concept contributions across all samples, revealing dominant, generalizable motion features.  
The results indicate that standing postures appear among the dominant concepts across the Penn Action dataset. 
This suggests that upright body positions serve as generalizable motion features, applicable across multiple actions in the dataset.
These findings highlight that \ours{} captures meaningful motion patterns not only at the sample level but also at broader scales, reinforcing its interpretability.
By analyzing concept contributions at both class and task levels, we gain deeper insights into how the model generalizes motion patterns, identifies shared features across actions, and differentiates between visually similar movements.




\section{Conclusion}
\label{sec:conclusions}

We introduce \ours{}, a novel framework for explainable human action recognition that uses human pose sequences as motion-aware concepts to provide interpretable and robust explanations. By explicitly modeling both spatial configurations and temporal dynamics, \ours{} overcomes the limitations of existing XAI methods based on static or pixel-level concepts.
Our extensive experiments across multiple datasets demonstrate that \ours{} outperforms baselines in both classification accuracy and interpretability. Additionally, \ours{} offers a significant advantage in explaining complex actions, making it highly valuable for real-world applications where model transparency is essential.
In future work, we plan to extend \ours{} to other domains and further optimize its scalability and efficiency, enhancing its broad applicability in the field of video understanding.

\clearpage
\paragraph{Acknowledgment.}
This work was supported by Institute of Information \& communications Technology Planning \& Evaluation (IITP) grant funded by the Korea government(MSIT) (RS-2019-II00330, Development of AI Technology for Early Screening of Infant/Child Autism Spectrum Disorders based on Cognition of the Psychological Behavior and Response, 50\%) and National Research Foundation of Korea (NRF) grant funded by the Korea Government (MSIT) (No. 2022R1F1A1070997, 50\%)
{\small
\bibliographystyle{ieee_fullname}
\bibliography{main}

\begin{thebibliography}{10}\itemsep=-1pt

\bibitem{slic}
Radhakrishna Achanta, Appu Shaji, Kevin Smith, Aurelien Lucchi, Pascal Fua, and Sabine S{\"u}sstrunk.
\newblock Slic superpixels compared to state-of-the-art superpixel methods.
\newblock {\em TPAMI}, 34(11):2274--2282, 2012.

\bibitem{achtibat2023attribution}
Reduan Achtibat, Maximilian Dreyer, Ilona Eisenbraun, Sebastian Bosse, Thomas Wiegand, Wojciech Samek, and Sebastian Lapuschkin.
\newblock From attribution maps to human-understandable explanations through concept relevance propagation.
\newblock {\em Nature Machine Intelligence}, 5(9):1006--1019, 2023.

\bibitem{adebayo2018sanity}
Julius Adebayo, Justin Gilmer, Michael Muelly, Ian Goodfellow, Moritz Hardt, and Been Kim.
\newblock Sanity checks for saliency maps.
\newblock In {\em NeurIPS}, 2018.

\bibitem{ahn2024unified}
Yong~Hyun Ahn, Hyeon~Bae Kim, and Seong~Tae Kim.
\newblock Www: a unified framework for explaining what where and why of neural networks by interpretation of neuron concepts.
\newblock In {\em CVPR}, 2024.

\bibitem{vivit}
Anurag Arnab, Mostafa Dehghani, Georg Heigold, Chen Sun, Mario Lu{\v{c}}i{\'c}, and Cordelia Schmid.
\newblock Vivit: A video vision transformer.
\newblock In {\em ICCV}, 2021.

\bibitem{DCR}
Pietro Barbiero, Gabriele Ciravegna, Francesco Giannini, Mateo~Espinosa Zarlenga, Lucie~Charlotte Magister, Alberto Tonda, Pietro Li{\'o}, Frederic Precioso, Mateja Jamnik, and Giuseppe Marra.
\newblock Interpretable neural-symbolic concept reasoning.
\newblock In {\em ICML}, 2023.

\bibitem{belharbi2023tcam}
Soufiane Belharbi, Ismail Ben~Ayed, Luke McCaffrey, and Eric Granger.
\newblock Tcam: Temporal class activation maps for object localization in weakly-labeled unconstrained videos.
\newblock In {\em WACV}, 2023.

\bibitem{timesformer}
Gedas Bertasius, Heng Wang, and Lorenzo Torresani.
\newblock Is space-time attention all you need for video understanding?
\newblock In {\em ICML}, 2021.

\bibitem{caetano2019skelemotion}
Carlos Caetano, Jessica Sena, Fran{\c{c}}ois Br{\'e}mond, Jefersson~A Dos~Santos, and William~Robson Schwartz.
\newblock Skelemotion: A new representation of skeleton joint sequences based on motion information for 3d action recognition.
\newblock In {\em IEEE international conference on advanced video and signal based surveillance}, 2019.

\bibitem{i3d}
Joao Carreira and Andrew Zisserman.
\newblock Quo vadis, action recognition? a new model and the kinetics dataset.
\newblock In {\em CVPR}, 2017.

\bibitem{ProtoPNet}
Chaofan Chen, Oscar Li, Daniel Tao, Alina Barnett, Cynthia Rudin, and Jonathan~K Su.
\newblock This looks like that: deep learning for interpretable image recognition.
\newblock In {\em NeurIPS}, 2019.

\bibitem{haa500}
Jihoon Chung, Cheng-hsin Wuu, Hsuan-ru Yang, Yu-Wing Tai, and Chi-Keung Tang.
\newblock Haa500: Human-centric atomic action dataset with curated videos.
\newblock In {\em ICCV}, 2021.

\bibitem{doshi2017towards}
Finale Doshi-Velez and Been Kim.
\newblock Towards a rigorous science of interpretable machine learning.
\newblock {\em arXiv preprint arXiv:1702.08608}, 2017.

\bibitem{vit}
Alexey Dosovitskiy, Lucas Beyer, Alexander Kolesnikov, Dirk Weissenborn, Xiaohua Zhai, Thomas Unterthiner, Mostafa Dehghani, Matthias Minderer, G Heigold, S Gelly, et~al.
\newblock An image is worth 16x16 words: Transformers for image recognition at scale.
\newblock In {\em ICLR}, 2020.

\bibitem{du2015hierarchical}
Yong Du, Wei Wang, and Liang Wang.
\newblock Hierarchical recurrent neural network for skeleton based action recognition.
\newblock In {\em CVPR}, 2015.

\bibitem{duan2022revisiting}
Haodong Duan, Yue Zhao, Kai Chen, Dahua Lin, and Bo Dai.
\newblock Revisiting skeleton-based action recognition.
\newblock In {\em CVPR}, 2022.

\bibitem{cem}
Mateo Espinosa~Zarlenga, Pietro Barbiero, Gabriele Ciravegna, Giuseppe Marra, Francesco Giannini, Michelangelo Diligenti, Zohreh Shams, Frederic Precioso, Stefano Melacci, Adrian Weller, Pietro Lio, and Mateja Jamnik.
\newblock Concept embedding models: Beyond the accuracy-explainability trade-off.
\newblock In {\em NeurIPS}, 2022.

\bibitem{x3d}
Christoph Feichtenhofer.
\newblock X3d: Expanding architectures for efficient video recognition.
\newblock In {\em CVPR}, 2020.

\bibitem{feichtenhofer2019slowfast}
Christoph Feichtenhofer, Haoqi Fan, Jitendra Malik, and Kaiming He.
\newblock Slowfast networks for video recognition.
\newblock In {\em ICCV}, 2019.

\bibitem{perturbation}
Ruth Fong, Mandela Patrick, and Andrea Vedaldi.
\newblock Understanding deep networks via extremal perturbations and smooth masks.
\newblock In {\em ICCV}, 2019.

\bibitem{fong2017interpretable}
Ruth~C Fong and Andrea Vedaldi.
\newblock Interpretable explanations of black boxes by meaningful perturbation.
\newblock In {\em ICCV}, 2017.

\bibitem{ace}
Amirata Ghorbani, James Wexler, James~Y Zou, and Been Kim.
\newblock Towards automatic concept-based explanations.
\newblock In {\em NeurIPS}, 2019.

\bibitem{hartley2022swag}
Thomas Hartley, Kirill Sidorov, Christopher Willis, and David Marshall.
\newblock Swag-v: explanations for video using superpixels weighted by average gradients.
\newblock In {\em WACV}, 2022.

\bibitem{hase2021out}
Peter Hase, Harry Xie, and Mohit Bansal.
\newblock The out-of-distribution problem in explainability and search methods for feature importance explanations.
\newblock In {\em NeurIPS}, 2021.

\bibitem{resnet}
Kaiming He, Xiangyu Zhang, Shaoqing Ren, and Jian Sun.
\newblock Deep residual learning for image recognition.
\newblock In {\em CVPR}, 2016.

\bibitem{jacovi2021formalizing}
Alon Jacovi, Ana Marasovi{\'c}, Tim Miller, and Yoav Goldberg.
\newblock Formalizing trust in artificial intelligence: Prerequisites, causes and goals of human trust in ai.
\newblock In {\em Proceedings of the 2021 ACM conference on fairness, accountability, and transparency}, 2021.

\bibitem{3d-ace}
Ying Ji, Yu Wang, and Jien Kato.
\newblock Spatial-temporal concept based explanation of 3d convnets.
\newblock In {\em CVPR}, 2023.

\bibitem{kinetics}
Will Kay, Joao Carreira, Karen Simonyan, Brian Zhang, Chloe Hillier, Sudheendra Vijayanarasimhan, Fabio Viola, Tim Green, Trevor Back, Paul Natsev, et~al.
\newblock The kinetics human action video dataset.
\newblock {\em arXiv preprint arXiv:1705.06950}, 2017.

\bibitem{tcav}
Been Kim, Martin Wattenberg, Justin Gilmer, Carrie Cai, James Wexler, Fernanda Viegas, et~al.
\newblock Interpretability beyond feature attribution: Quantitative testing with concept activation vectors (tcav).
\newblock In {\em ICML}, 2018.

\bibitem{cbm}
Pang~Wei Koh, Thao Nguyen, Yew~Siang Tang, Stephen Mussmann, Emma Pierson, Been Kim, and Percy Liang.
\newblock Concept bottleneck models.
\newblock In {\em ICML}, 2020.

\bibitem{vcad}
Matthew Kowal, Achal Dave, Rares Ambrus, Adrien Gaidon, Konstantinos~G Derpanis, and Pavel Tokmakov.
\newblock Understanding video transformers via universal concept discovery.
\newblock In {\em CVPR}, 2024.

\bibitem{cast}
Dongho Lee, Jongseo Lee, and Jinwoo Choi.
\newblock Cast: Cross-attention in space and time for video action recognition.
\newblock In {\em NeurIPS}, 2023.

\bibitem{video-perturbation}
Zhenqiang Li, Weimin Wang, Zuoyue Li, Yifei Huang, and Yoichi Sato.
\newblock Towards visually explaining video understanding networks with perturbation.
\newblock In {\em WACV}, 2021.

\bibitem{coco}
Tsung-Yi Lin, Michael Maire, Serge Belongie, James Hays, Pietro Perona, Deva Ramanan, Piotr Doll{\'a}r, and C~Lawrence Zitnick.
\newblock Microsoft coco: Common objects in context.
\newblock In {\em ECCV}, 2014.

\bibitem{SHAP}
Scott~M Lundberg and Su-In Lee.
\newblock A unified approach to interpreting model predictions.
\newblock In {\em NeurIPS}, 2017.

\bibitem{lrp}
Gr{\'e}goire Montavon, Alexander Binder, Sebastian Lapuschkin, Wojciech Samek, and Klaus-Robert M{\"u}ller.
\newblock Layer-wise relevance propagation: an overview.
\newblock {\em Explainable AI: interpreting, explaining and visualizing deep learning}, pages 193--209, 2019.

\bibitem{lf-cbm}
Tuomas Oikarinen, Subhro Das, Lam Nguyen, and Lily Weng.
\newblock Label-free concept bottleneck models.
\newblock In {\em ICLR}, 2023.

\bibitem{Dissect}
Tuomas Oikarinen and Tsui-Wei Weng.
\newblock Clip-dissect: Automatic description of neuron representations in deep vision networks.
\newblock In {\em ICLR}, 2023.

\bibitem{motionformer}
Mandela Patrick, Dylan Campbell, Yuki Asano, Ishan Misra, Florian Metze, Christoph Feichtenhofer, Andrea Vedaldi, and Joao~F Henriques.
\newblock Keeping your eye on the ball: Trajectory attention in video transformers.
\newblock In {\em NeurIPS}, 2021.

\bibitem{ta-cbm}
Sukrut Rao, Sweta Mahajan, Moritz Böhle, and Bernt Schiele.
\newblock Discover-then-name: Task-agnostic concept bottlenecks via automated concept discovery.
\newblock In {\em ECCV}, 2024.

\bibitem{lime}
Marco~Tulio Ribeiro, Sameer Singh, and Carlos Guestrin.
\newblock Why should i trust you? explaining the predictions of any classifier.
\newblock In {\em Proceedings of the 22nd ACM SIGKDD international conference on knowledge discovery and data mining}, 2016.

\bibitem{video-tcav}
Avinab Saha, Shashank Gupta, Sravan~Kumar Ankireddy, Karl Chahine, and Joydeep Ghosh.
\newblock Exploring explainability in video action recognition.
\newblock In {\em CVPR Workshop}, 2024.

\bibitem{finch}
Saquib Sarfraz, Vivek Sharma, and Rainer Stiefelhagen.
\newblock Efficient parameter-free clustering using first neighbor relations.
\newblock In {\em CVPR}, 2019.

\bibitem{kth}
C. Schuldt, I. Laptev, and B. Caputo.
\newblock Recognizing human actions: a local svm approach.
\newblock In {\em ICPR}, 2004.

\bibitem{gradcam}
Ramprasaath~R Selvaraju, Michael Cogswell, Abhishek Das, Ramakrishna Vedantam, Devi Parikh, and Dhruv Batra.
\newblock Grad-cam: Visual explanations from deep networks via gradient-based localization.
\newblock In {\em ICCV}, 2017.

\bibitem{increcbm}
Chenming Shang, Shiji Zhou, Hengyuan Zhang, Xinzhe Ni, Yujiu Yang, and Yuwang Wang.
\newblock Incremental residual concept bottleneck models.
\newblock In {\em CVPR}, 2024.

\bibitem{shi2019skeleton}
Lei Shi, Yifan Zhang, Jian Cheng, and Hanqing Lu.
\newblock Skeleton-based action recognition with directed graph neural networks.
\newblock In {\em CVPR}, 2019.

\bibitem{slack2020fooling}
Dylan Slack, Sophie Hilgard, Emily Jia, Sameer Singh, and Himabindu Lakkaraju.
\newblock Fooling lime and shap: Adversarial attacks on post hoc explanation methods.
\newblock In {\em Proceedings of the AAAI/ACM Conference on AI, Ethics, and Society}, 2020.

\bibitem{googlenet}
Christian Szegedy, Wei Liu, Yangqing Jia, Pierre Sermanet, Scott Reed, Dragomir Anguelov, Dumitru Erhan, Vincent Vanhoucke, and Andrew Rabinovich.
\newblock Going deeper with convolutions.
\newblock In {\em CVPR}, 2015.

\bibitem{craft}
Fel Thomas, Picard Agustin, Bethune Louis, Boissin Thibaut, Vigouroux David, Colin Julien, Cadène Rémi, and Serre Thomas.
\newblock Craft: Concept recursive activation factorization for explainability.
\newblock In {\em CVPR}, 2023.

\bibitem{videomae}
Zhan Tong, Yibing Song, Jue Wang, and Limin Wang.
\newblock Video{MAE}: Masked autoencoders are data-efficient learners for self-supervised video pre-training.
\newblock In {\em NeurIPS}, 2022.

\bibitem{c3d}
Du Tran, Lubomir Bourdev, Rob Fergus, Lorenzo Torresani, and Manohar Paluri.
\newblock {Learning spatiotemporal features with 3d convolutional networks}.
\newblock In {\em ICCV}, 2015.

\bibitem{occlusion_video}
Tomoki Uchiyama, Naoya Sogi, Koichiro Niinuma, and Kazuhiro Fukui.
\newblock Visually explaining 3d-cnn predictions for video classification with an adaptive occlusion sensitivity analysis.
\newblock In {\em WACV}, 2023.

\bibitem{sparselinear}
Eric Wong, Shibani Santurkar, and Aleksander Madry.
\newblock Leveraging sparse linear layers for debuggable deep networks.
\newblock In {\em ICML}, 2021.

\bibitem{pose_estimator}
Yufei Xu, Jing Zhang, Qiming Zhang, and Dacheng Tao.
\newblock Vitpose: Simple vision transformer baselines for human pose estimation.
\newblock In {\em NeurIPS}, 2022.

\bibitem{vitpose}
Yufei Xu, Jing Zhang, Qiming Zhang, and Dacheng Tao.
\newblock Vitpose: Simple vision transformer baselines for human pose estimation.
\newblock In {\em NeurIPS}, 2022.

\bibitem{stgcn}
Sijie Yan, Yuanjun Xiong, and Dahua Lin.
\newblock Spatial temporal graph convolutional networks for skeleton-based action recognition.
\newblock In {\em AAAI}, 2018.

\bibitem{pcbm}
Mert Yuksekgonul, Maggie Wang, and James Zou.
\newblock Post-hoc concept bottleneck models.
\newblock In {\em ICLR}, 2023.

\bibitem{zeiler2014visualizing}
Matthew~D Zeiler and Rob Fergus.
\newblock Visualizing and understanding convolutional networks.
\newblock In {\em ECCV}, 2014.

\bibitem{penn}
Weiyu Zhang, Menglong Zhu, and Konstantinos~G. Derpanis.
\newblock From actemes to action: A strongly-supervised representation for detailed action understanding.
\newblock In {\em ICCV}, 2013.

\bibitem{cam}
Bolei Zhou, Aditya Khosla, Agata Lapedriza, Aude Oliva, and Antonio Torralba.
\newblock Learning deep features for discriminative localization.
\newblock In {\em CVPR}, 2016.

\bibitem{ibd}
Bolei* Zhou, Yiyou* Sun, David* Bau, and Antonio Torralba.
\newblock Interpretable basis decomposition for visual explanation.
\newblock In {\em ECCV}, 2018.

\end{thebibliography}
}

\newpage
\appendix
\section*{Appendix}





\section{Details of datasets.}
\label{appen:dataset}
\paragraph{KTH-5}
The KTH~\cite{kth} dataset consists of 25 actors, each performing six actions across four different environments. The originally proposed splits ensure that videos of a single actor remain within one set, as the dataset was not designed for identity recognition.

\paragraph{Penn Action}
The Penn Action~\cite{penn} dataset includes 2,326 video sequences covering 15 different actions with detailed human joint annotations. Each frame is stored in RGB format with a resolution of up to 640×480 pixels. The dataset provides 2D joint locations, visibility annotations, bounding boxes, coarse viewpoint labels, and train/test splits, making it well-suited for action recognition and pose estimation tasks.
\paragraph{HAA49} 
The HAA500~\cite{haa500} dataset consists of 500 fine-grained action classes with over 591K labeled frames, focusing on precise human-centric atomic actions. Unlike coarse-grained datasets, it ensures clear class distinctions (e.g., “Baseball Pitching” vs. “Free Throw in Basketball”) and maintains high pose detectability. Carefully curated to eliminate irrelevant motions and label noise, HAA500 is well-suited for detailed action recognition and human pose analysis.
The HAA49 dataset is a subset of HAA500, consisting of 49 action classes that focus on fine-grained, human-centric actions. 
We use the following classes from the HAA49 dataset for sampling: \texttt{yoga\_bridge}, \texttt{yoga\_cat}, \texttt{yoga\_dancer}, \texttt{yoga\_firefly}, \texttt{yoga\_fish}, \texttt{yoga\_gate}, \texttt{yoga\_locust}, \texttt{yoga\_lotus}, \texttt{yoga\_pigeon}, \texttt{yoga\_tree}, \texttt{yoga\_triangle}, \texttt{yoga\_updog}, \texttt{weightlifting\_hang}, \texttt{weightlifting\_overhead}, \texttt{weightlifting\_stand}, \texttt{volleyball\_overhand}, \texttt{volleyball\_pass}, \texttt{volleyball\_set}, \texttt{volleyball\_underhand}, \texttt{tennis\_serve}, \texttt{taekwondo\_middle\_block}, \texttt{sprint\_start}, \texttt{soccer\_shoot}, \texttt{soccer\_throw}, \texttt{situp}, \texttt{pushup}, \texttt{punching\_sandbag}, \texttt{pull\_ups}, \texttt{one\_arm\_push\_up}, \texttt{leg\_hold\_back}, \texttt{leg\_hold\_front}, \texttt{leg\_split}, \texttt{high\_knees}, \texttt{handstand}, \texttt{gym\_squat}, \texttt{gym\_run}, \texttt{gym\_ride}, \texttt{gym\_plank}, \texttt{gym\_pull}, \texttt{gym\_push}, \texttt{gym\_lunges}, \texttt{gym\_lift}, \texttt{golf\_swing}, \texttt{dips}, \texttt{bench\_dip}, \texttt{baseball\_swing}, \texttt{backflip}, \texttt{arm\_wave}, and \texttt{jumping\_jack}.
These carefully curated classes in HAA49 enable detailed analysis of pose-based actions and movements, making it an excellent resource for video action recognition tasks focused on human poses and dynamic motion.

\section{Details of implementations.}
\label{appen:implementations}
We implement our framework using PyTorch, built upon the Label-Free CBM~\cite{lf-cbm} code base. The framework utilizes a Nvidia RTX 3090 GPU for training and evaluation.
For feature extraction, we use VideoMAE-B/16~\cite{videomae}, pretrained on Kinetics-400~\cite{kinetics}. 
For pose estimation, we employ the off-the-shelf ViTPose-B~\cite{vitpose} model, pretrained on COCO~\cite{coco}, and leverage 17 keypoints from COCO for pose extraction.
To create the concept set for the textual concept baseline, as shown in Figure 1 and Table 1 of the main paper, we follow the Label-Free-CBM~\cite{lf-cbm} method for selecting and defining concepts. The Label-Free-CBM provides the code base for concept set processor, which we use to generate concepts tailored to each dataset. To match the number of concepts with the number of classes, we remove concepts that are similar to class names.
We train the concept layer with a batch size of 256 for 1,000 steps. The sparse linear layer is optimized with a batch size of 512, a learning rate of 1e-3, $\alpha = 0.99$, and a step size of 0.05 for controlled updates.
For input clips, we set the frame length $l = 16$ frames per clip, and each pose is extracted using the 17 keypoints from the MS COCO~\cite{coco} dataset, with $J = 17$ keypoints for each frame.
These configurations ensure that the model efficiently learns both spatial and temporal dynamics, providing high-quality representations for action recognition while maintaining interpretability and robustness.

\section{Metric.}
\label{appen:metric}

To adapt the Concept Utilization Efficiency (CUE) metric from Res-CBM~\cite{increcbm} for our framework, we refer to the original formulation:

\[
CUE = \dfrac{10000 \times \text{Acc}}{N \times \bar{L}},
\]
where \( \text{Acc} \) represents the classification accuracy, \( N \) is the number of concepts, and \( \bar{L} \) is the average length of the concepts.

In Res-CBM, the multiplication by 10,000 is used to scale the metric, allowing it to measure the improvement in classification accuracy considering both the length of the concepts and the total number of concepts. Larger values of CUE indicate higher efficiency in utilizing concepts.
For our case, since we are working with pose-based concepts derived from skeleton sequences, we cannot define concept length in the same way as textual concepts in Res-CBM. Therefore, we set \( \bar{L} = 1 \), as each pose concept corresponds to a fixed-length representation (such as a specific pose or sequence of poses) rather than a variable-length textual description.
Additionally, to avoid an overly large scale due to multiplying by 10,000, we use a more manageable scaling factor of 100. This adjustment ensures the CUE score remains interpretable while still reflecting the efficiency of concept utilization.
Thus, our modified CUE metric becomes:
\[
CUE^* = \dfrac{100 \times \text{Acc}}{N}.
\]
This formula reflects the efficiency of concept utilization in our framework, where \( \text{Acc} \) is the top-1 accuracy, and \( N \) is the number of concepts. This modification enables us to assess the effectiveness of pose-based concepts in video action recognition tasks, with a focus on maintaining simplicity and clarity in the calculation of CUE.

Additionally, to evaluate the effectiveness of our clustering results, we employ the Normalized Mutual Information (NMI) score, which quantifies the alignment between the learned concept clusters and ground truth labels. 
It computes how much information is shared between the two distributions: the learned clusters and the true labels. A higher NMI score indicates that the clusters are more consistent with the true class labels, reflecting a more accurate clustering of the concepts. 
NMI is normalized to ensure that the score is independent of the number of clusters, making it a robust and interpretable metric for evaluating the quality of the concept clusters.

\end{document}


\title{PCBEAR: Pose Concept Bottleneck for Explainable Action Recognition}

\maketitlesupplementary

\input{99_supp_content}
{
\small
\bibliographystyle{ieee_fullname}
\bibliography{main}
}